\documentclass{article}
\usepackage[ansinew]{inputenc} 
\usepackage{graphicx}
\usepackage{amsmath}
\usepackage{amssymb}

\usepackage{color}

\usepackage[labelfont=bf,labelsep=period,justification=raggedright]{caption}

\makeatletter
\renewcommand{\@biblabel}[1]{\quad#1.}
\makeatother

\date{}

\begin{document}
\title{Optimal choice: new machine learning problem and its solution}

\author{Marina Sapir\\
metaPattern\\
http://sapir.us
}

\maketitle

\section* {Abstract}

The task of  learning to pick a single preferred example out a finite set of examples, an ``optimal choice problem'', is a supervised machine learning problem  with complex, structured input.  Problems of optimal choice emerge  often in various practical applications.  We formalize the problem, show that it does not satisfy the assumptions of statistical learning theory, yet it can be solved efficiently in some cases. We  propose two approaches to solve the problem. Both of them reach good solutions on real life data from a signal processing application.

\section{Introduction}

Machine learning is described as  `automated detection of meaningful patterns in data' \cite{Shalev}. No limitations on types of data or patterns are set explicitly.    Yet, supervised machine learning includes, mostly, three types of problems: regression, classification and ranking which all have  the \textit{relational form}:

\begin{enumerate}
\item \textit{Given} is a finite set of pairs $\{ \langle x_i, y_i \rangle \}$, where observations $x_i \in \mathbb{R}^n$, and $y_i$ are some labels,
\item  \textit{The goal} is  to find a function which  approximates dependence of labels  $y$ on vectors $x$.

\end{enumerate}

The problems differ by types of labels and measures of success but not by their general structure. The concepts ``learning models'', `PAC learnable'' in statistical learning theory are developed for the relational data \cite{Shalev, Hastie}.

Recently, the concept of ``structural output learning'' is being developed. There are several specific applied problems being considered (image and string segmentation, string labeling and alignments), but very little general formalizations. Yet, one may observe that in these problems, the inputs $x$ are, usually, reduced to a vectors of certain length, but the labels may have a complex structure and be represented by sequences of different, arbitrary lengths \cite{Ricci, Deep}.

There is unstated assumption that any machine learning problem can be presented in relational form above.

However, in some real life learning problems,  inputs can not be represented by vectors of fixed lengths.  Consider, for example the  task of predicting an outcome of a horse race.

Given is the history of horse races,  including the information about the participating horses and the  winner of each race. Each competing horse is characterized by certain features. The goal is, knowing the features of the horses running in the next race, predict the winner.

This is a supervised machine learning problem, since it requires to automatically learn prediction of the winning horse. Yet, the problem is not equivalent  to any relational form problem. The main issue here is that  victory depends not only on  qualities of the winner, but on qualities of other competing horses in the same race. A horse may be sure winner in one race and sure loser in another race, where all the horses are ``better'' in some important ways. As races are different, so are the winners. The data reflect relationship not between horses and the victories, but between horses and races on one hand,  and the victories on another. Since a race may have arbitrary number of running horses, the description of the race can not be presented as a single vector of fixed length.

Here, we will study the general problem which requires to learn how to pick the ``best'' example out of finite sets of examples. We call it an \textit{Optimal Choice} (OC ) problem, of which horse races is one example. We will start by giving  the formal definition. Then we bring examples of important and popular  applications where such problems appear. We show that the main assumptions of statistical learning theory are violated for OC. Next, we explore two general approaches to solve the problem and test them on a real life data concerning finding the true cycle out of several candidate intervals in a practical signal processing application.

\subsection{Optimal Choice Problem, Formal Statement}

Let us introduce the general concepts and the definition of the problem.

A \textit{choice} is vector of values of $n$ features, $x  \subset \Re^n$. A \textit{lot}  is a finite set of choices. Denote $D$ set of possible choices, and  $\Omega$  set  of all possible lots.

Each lot has not more  than  one choice identified as \textit{prime}. Conveniently denote $X^\prime$ the prime in the lot $X$.  If the lot $X$ does not have a prime, it will be denoted as $X^\prime = \emptyset.$

Then,  training set can be presented as a finite set of pairs $$    Z = \{ \langle X_1, I(X_1^\prime, X_1)  \rangle, ..., \langle X_m, I(X_m ^\prime, X_m) \rangle \},$$ where all $X_i \in \Omega$ and  $I(x, X)$ is identity function on $X:$

\begin{enumerate}
\item $I(x, X) \in \{0, 1\}$
\item $I(x, X) = 1 \Leftrightarrow x = X^\prime$
\item if $X^\prime = \emptyset, I(x, X) \equiv 0.$
\end{enumerate}

For example, in horse races, \textit{choices} are the horses, a \textit{lot} is a race, and the \textit{prime} is the winner of the race.

The goal is to build a \textit{labeling function} $f(x, X) \rightarrow \{0, 1\}$ such that, for every $ X \in \Omega, x \in X, f(x, X) = I(x, X).$

If the condition $f(x, X) = I(x, X)$ holds for a labeling function $f$ and lot $X$, we say that a lot $X$ is a \textit{success } of a labeling function $f.$
 \textit{Success rate} of a labeling function is the probability of its successes on $\Omega.$

Let us consider a numeric function $g(x, X)$ on pairs $X \in \Omega, x \in X.$
The function $g$ correspond to a single labeling function
$$f(x, X) = 1 \Leftrightarrow \arg \, \max_{x \in X} g(x, X) = x$$ if $g(x, X)$ has a single maximum on $X;$ otherwise,
$f(x, X) \equiv \emptyset$ on $X.$

The function $g(x, X)$ which identifies the primes by its maxima in the lots will be called \textit{scoring function}. The success rate of the corresponding labeling function will be associated with the scoring function.

\subsection{Practical examples of the optimal choice problem}

OC problem occurs in many practical applications, even though it was not formalized yet, to the best of author's knowledge.
\begin {enumerate}

\item \textbf{Competitions}

Not only horse races, but most of competitions,   including sports, beauty contests, elections,  tender awards and so on, give rice to the OC type problems.  Every competition with a single winner has the same form of data, essentially.   Even when the same participants compete in each round of the competition, they change with time and their features change as well.

\item \textbf{Finding cycle in continuous noisy signal}

  The OC  type problem was discovered in real life data analysis as part of the large signal processing research. The company Predictive Fleet Technology, for which the author did consulting,  analyzes signal from piezoelectric sensors, installed in vehicles. The engine's `signature' (recorded signal) continuously reflects changes in pressure in exhaust pipe and crankcase, which occur when engine works. The cylinders in an engine fire consequently, so it  should be possible to identify intervals of work of each cylinder within the signature. The goal is to evaluate the regularity of the engine and identify possible issues.

  The most important part of the signature interpretation is to find the \textit{cycle}: interval of time, when every cylinder works once. The problem is difficult when the signature is irregular and curves of consecutive cycles do not look the same, and when the signature is very regular and all the cylinders look identical. Some preliminary work allowed us to find the intervals of potential cycles (choices) for each signature, and each choice is characterized by four ``quality criteria'' (features).  There is a training set, where an expert identified true cycle for each signature.

   The features are obtained by aggregating several signal characteristics: two features evaluate irregularity of each of the curves (from exhaust pipe and crankcase) would have if the given interval is selected as the cycle. And two other features characterize some measures of complexity of the interval itself. All the features correlate negatively with the likelihood of the choice being the prime. Three features are continuous, and the fourth feature is binary. They are scaled from 0 to 1.

   The goal is to develop the rule which identifies the true cycle (the prime) among the chosen intervals for each signature. The main property of this problem is that there is no ``second best'': only one cycle is correct, the rest are equally wrong. This problem lands naturally into the general OC problem.

   All the solutions proposed here are applied on the dataset of this problem.  The data contain 2453 choices in 114 lots, on average $~21$ choice per lot, no lot contains less than 2 choices, and every lot has a prime.

   The data are available upon request.
   
   \item \textbf{Image and signal segmentation}
   
   The problem of finding a cycle is an example from a large class of problems of image and signal segmentation. Suppose, a preprocessing algorithm can identify variants of segmentation for a particular image. If there are some criteria which can characterize each segmentation, then a \textit{lot} will consist of the the variants of the segmentation for the given image with the features given by criteria values. If an expert marks a correct segmentation for each image, we are in the situation of OC problem again.

  \item \textbf{Multiple classes classification with binary classifiers}

  In many cases,  the classification with $k > 2$ classes is done with a binary classifier. The task is split on classifications for each one class versus all others. But what if some instances do not fit nicely in any of the classes, or found similar to more than one class? What if one uses several classification methods which point to different classes? One still needs to find the ``prime'' class. In these types of problems, each lot will have $k$ choices, and each choice will be characterized by some criteria of fit between the class and the instance. One needs to find an optimal rule which will aggregate these single class criteria into a rule for all the classes together. This is an OC problem.

\item \textbf{Recommender system}
    Suppose, a  recommendation system presents a customer with sets of choices each time, and lets him to choose one option he likes the best. The choices  may be movies, books, real estate, fashions and so on.  The goal is  to learn from the customer's past choices and recommend him new choices in the order of his preferences.  Here, we are in situation of the OC problem again. The training set contains past lots, where each choice is characterized by several features, and the customer's choice (prime) is known.   The system needs to develop a personalized scoring function on choices to present them to the customer in order of his preferences.

\item \textbf{Rating system}
Let us consider a rating  problem. There are two types of such problems, depending on the feedback. The feedback in the training set may be binary, or it may represent ranks. For example, the trainer marks each link as ``relevant'' to a query  or not. Another option is to have the trainer to assign a rank (or rating) to each object in the training set. In both cases, the goal is to learn to rank the new objects.

Both approaches are hard on the trainer. If actual ratings have many values (various degrees of relevancy), it may be difficult to assign just two values correctly. Assigning multiple rating in the training set may be even more difficult.  For a normal size training set, it would be too much work for a trainer to check all the comparisons his ratings imply. Besides, some of the objects he rates may not be comparable. It means, the trainer can not guaranty that all the relationships implied by his ranking are true. It leads to inevitable errors, contradictions in labels in training data.

 The solution may be to ask the trainer to select  groups of comparable objects (lots), and identify the best choice (prime)  in each group.  This will lead one to OC problem.

 \end{enumerate}

 Finding robust solutions for any of these practical applications may be very valuable. Yet, it will require some new approaches.

 \subsection{OC and statistical learning theory}
\label{SLT}

The problem has some similarity with two popular types of machine learning problems:  classification and ranking.

As in binary classification, the goal in OC is to learn a rule, which can be applied on the new data to classify each choice in the lot as its prime or not.

To see similarity with the ranking problem, let us notice that selecting the prime in each lot establishes partial order on the set of choices of this lot: $X^\prime \succ q$ for a $q \in X, q \neq X^\prime.$ So, the OC problem can be considered as a problem of learning the partial order on the samples of the order.

Despite the similarities, the problem can not be presented in relational form, and the main explicit and implicit assumptions of statistical learning theory do not hold  for OC.

\begin{enumerate}

\item \textbf{Labels are not the function of the features}

Statistical learning  implicitly assumes that the values of the features determine  the probability of a label. Essentially, each classification method approximates a random function on $D \rightarrow \mathbb{R}$ given on the training set.

   In OC, the labels are not a function of the features alone, since they depend both on the choice and its lot.   A choice may be the prime in one lot and not in another. For example, the prime horse in a small stable is expected to be a poor competitor in some famous derby. It means that close (or even identical) choices in different lots shall be assigned different labels by the learning algorithm to have satisfactory success rate.

\item \textbf{The fit can not be evaluated point-wise}

In classical machine learning,  the  fit between the true labels  and assigned labels is estimated as a measure of success, accuracy. For example, in classification, the probability of the correct labels is evaluated. In ranking, some measure of the correlation (agreement in order) between the known ranks and predicted scores is estimated.

The next examples show that counting correct labels on choices or measuring correlation between the assigned and true labels in the training set can not be used to evaluate the learning success in  OC.

Suppose, for example, there are 10 choices in every lot. From classification point of view, if the decision rule assigns zero to every choice, the rule is $90\%$ correct. From optimal choice  point of view,  the success rate is 0, because it did not identify any of the  primes.

If a scoring function scores a prime in every lot as the ``second-best'', many correlation measures used in supervised ranking will be rather high.  In this case, on each lot, 8 out of every 9 not-prime choices are below the only prime choice, so  AUC = 8/9 \cite{Rudin}.  Yet, the scoring function fails to find  the prime everywhere, and, accordingly,  the success rate of this scoring function is 0.

\item \textbf{Independence}

In machine learning, both feature vectors and feedback are supposed to be taken independently from the same distribution. This is, obviously, not the case here.

As for labels, in each lot, only one label is 1, the rest are 0.

The distribution of the choice features is expected to depend on the  lot.  For example,  more prestigious races will include better overall participants and have necessarily different from other races distribution of features.

However, we can expect that the lots appear independently, according with some probability distribution $Pr(X)$ on $\Omega$. Also, we can assume that there is probability distribution of a choice to be the prime in a given lot: $Q(x, X) = Pr(x = X^\prime). $ The lots and their primes in the training set are generated in accordance with these two distributions.

\end{enumerate}

\section{Solving the problem }

The statistical learning assumptions make classic machine learning problems tractable, allow some efficient solutions. We showed that the assumptions are not satisfied in OC type of problems. So, one may wonder, if OC has a decent solution.

In fact, identifying these issues helps us to find the ways the problem can be solved. We explore two paths to the solution here.

First, we consider the problem in an extended set of features, where the added features characterize lots. If the features are selected successfully, it can make the labels dependent on the choices only. Then, the point-wise fitting machine learning methods can be applied, provided that, in the end, the fit is still evaluated with the success rate.

As another possible solution, we explore general optimization methods which can optimize the OC success rate directly in original feature space.

We will show here on the example of the real life and difficult problem of finding cycle in continuous noisy signal that a good solution can be found efficiently both ways.

\subsection{Expanding  feature space to apply popular machine learning methods}

  As we mentioned above, the main issue with OC problem is that the labels depend both on the choice and its lot. To make the  labels less dependent of lots, we add new features about lot as a whole. Denote $\overline{D}$ the extended feature space. In $\overline{D}$, each choice still has its own specific features as well as new features, characterizing its lot,  and common for every choice in its lot.

   The goal of extending the feature space is to have similar in  $\overline{D}$ choices  across all lots to have identical labels with high likelihood.

  Selection of the lot features, usually, requires some domain knowledge. However, there may be some empiric considerations which simplify the selection.

     Let us consider a simple case, when  there are features $F$ which correlate with the likelihood of a choice to be prime and mutually correlate (if the features $F$ are developed to predict primes, it is the case, usually).  Denote $m_X$ vector of maximal values of the features $F$ in lot $X$, and suppose values of  $m_X$ are  used as new features to characterize the lot. Then, a lot with higher values of  $m_X$ will, likely, have a prime with higher values of the features $F$. It is likely as well that lots with similar values of $m_X$ will have similar primes. Or, at the very least, their primes will be less dissimilar, than primes of the lots with very different features $m_X$.

 In \textit{finding the cycle} problem,  two features, cc.d and CrankRegul,  have very different distributions from one engine to another because the regularity of engines varies widely. The features have negative correlation with the likelihood of the choice to be prime. So  we added two new features: \textit{min.cc.d}, and \textit{min.CrankRegul}, which equal minimal in the lot values of the features \textit{cc.d }and \textit{CrankRegul} respectively.

 For good engines, the features correlate strongly. They both achieve minima on the intervals multiple of the true cycle.  For bad engines, the features do not correlate. It means, for a bad engine, there may be potential cycles with low value of one feature and high value of another. Then, the bad engine has low values of both additional features, as do  good engines. In this case, additional features do not help distinguishing the engines and predicting the prime. Fortunately, this does not happen often. The bad engines have, usually, higher values of these features than good engines.

With all the applied methods, the output was interpreted as a scoring function: primes were identified by maximal values of the function in each lot.

 We used the R implementation of the most popular regression  and classification methods: function \textit{SVM} with linear kernel from \textit{e1071} R package, the function \textit{neuralnet} from the R package with the same name, function \textit{boosting} from the R package \textit{adabag}, function \textit{glm} from R base to build logistic regression. In all the functions, we used predicted continuous output as a scoring function.

  Testing was done with ``leave one lot out'' procedure: each lot, consequently,  was removed from training and used for test. Percentages of the test lots, where the prime was correctly found  by each method,   are in the table \ref{ML}.

\begin{table}
  \centering
  \caption{Success Rate, Machine Learning Methods}\label{ML}
  $
\begin{array}{|l|l|l| }
\hline
  Method &  Orig. \; data & Extended \; data   \\\hline

  SVM, linear \; kernel   & 0.80 & 0.86 \\
  ADA boost & 0.74 & 0.83 \\
  Logistic \; regression & 0.82 & 0.87 \\
  Neural \; net, 3 \; neurons  & 0.79  & 0.79\\
  Neural \; net, 4 \; neurons  & 0.80  & 0.81\\ \hline

\end{array}
$

\end{table}

Selection of parameters of the neural net is a challenge. We used 1 hidden layer with 3  and 4 neurons. Adding more layers or neurons does not help in our experiments. ADAboost strongly depends on the selected method's parameters as well. In our experiments, ADAboost builds 30 decision trees with the depth not more than 5. Changing these parameters did not improve the solution.

\subsection{Optimization of the success rate criterion}

We were looking for a linear function of the original features which maximizes the success rate. The dataset at hand is relatively small, so we could use  the most general (slow) optimization methods, which do not use derivatives.

\subsubsection{Using standard optimization methods}

We applied the standard  Nelder - Mead  derivative-free optimization method (implemented in the function \textit{optim} of the R package \textit{stats}) to find the a linear function of the features optimizing the success rate criterion. Depending on the starting point, the success rate of the found functions on the whole dataset was from 0.82 to 0.86.  Other optimization methods implemented in the same R function produced worse results. It is interesting to notice that the the method works amazingly fast, much faster than neural network or ADAboost on the same data.

\subsubsection{Brute Force Optimization}

The data for the \textit{cycle} problem are rather small, so we applied the exhaustive search to find true optimal hypothesis in a narrow class of hypotheses defined by the next rules:
\begin{enumerate}
\item The hypotheses are linear functions of the features $f(x) = a_1 \cdot x_1 + \ldots + a_4 \cdot x_4$.
\item All coefficients $a_1, \ldots, a_4$ have integer values  from $0$ to $n$,
\end{enumerate}
where $n$ is a parameter of the algorithm. Out of all the possible rules with the same threshold $n$ and identical ($1\%$ tolerance) performance, the algorithms picks the rule with minimal sum of coefficients.

For $n = 15$,  the optimal linear function has the maximal value of coefficients equal 5. The rule has success rate $88\%$. The algorithm with the parameter $n = 5$ was applied in ``leave one lot out'' experiment with the same success rate, $88\%$.

\section{Conclusion}

Here we formulated the new machine learning problem, Optimal Choice, and proposed two paths to solve it. The problem emerges in various practical applications quite often, but it has undesirable properties from statistical learning theory point of view.  Formalization of the problem and close look at its statistical properties helped to find ways to solve it. The solutions were applied to a real life problem of signal processing with rather satisfactory results.

The proposed solutions have some shortcomings. The first approach is based on extending feature space, which would require some domain knowledge in each case. The direct optimization of the success rate criterion worked very well on relatively small data, but it may not be scalable.

The goal of this article will be achieved, if the OC problem attracts some attention and further research.

\bibliography{OptimalChoice}
\end{document}